\documentclass[conference]{IEEEtran}
\IEEEoverridecommandlockouts
\usepackage{cite}
\usepackage{amsmath,amssymb,amsfonts}
\usepackage{multirow}
\usepackage{algorithmic}
\usepackage{graphicx}
\usepackage{textcomp}
\usepackage{comment}
\usepackage{subcaption}
\usepackage{booktabs, multirow} 
\usepackage{soul}
\usepackage[table]{xcolor} 
\usepackage{changepage, threeparttable} 
\def\BibTeX{{\rm B\kern-.05em{\sc i\kern-.025em b}\kern-.08em
    T\kern-.1667em\lower.7ex\hbox{E}\kern-.125emX}}

\begin{document}

\title{Sensitivity Analysis of RF+clust for Leave-one-problem-out Performance Prediction

\thanks{Ana Nikolikj (Email: ana.nikolikj@ijs.si), Peter Koro\v{s}ec (Email: peter.korosec@ijs.si) and Tome Eftimov (Email: tome.eftimov@ijs.si) are with Computer Systems Department, Jo\v{z}ef Stefan Institute, 1000 Ljubljana, Slovenia. Ana Nikolikj is also with the Jo\v{z}ef Stefan International Postgraduate School, 1000 Ljubljana, Slovenia}
\thanks{Michal Pluh\'{a}\v{c}ek (Email:pluhacek@utb.cz) is with the Faculty of Applied Informatics, Tomas Bata University in Zlin, 
760 01 Zlin, Czech Republic.}
\thanks{Carola Doerr (Email: carola.doerr@lip6.fr) is with the Sorbonne Université, CNRS, LIP
1000 Paris, France.}
\thanks{The authors acknowledge the support of the Slovenian Research Agency through program grant No. P2-0098, project grants N2-0239 and J2-4460, and a bilateral project between Slovenia and France grant No. BI-FR/23-24-PROTEUS-001 (PR-12040). Our work is also supported by ANR-22-ERCS-0003-01 project VARIATION.} 
}

\author{\IEEEauthorblockN{Ana Nikolikj, Michal Pluh\'{a}\v{c}ek, Carola Doerr, Peter Koro\v{s}ec, 
and Tome Eftimov
      }
}


\maketitle
\IEEEpubidadjcol

\begin{abstract} 
Leave-one-problem-out (LOPO) performance prediction requires machine learning (ML) models to extrapolate algorithms' performance from a set of training problems to a previously unseen problem. LOPO is a very challenging task even for state-of-the-art approaches. Models that work well in the easier leave-one-instance-out scenario often fail to generalize well to the LOPO setting. To address the LOPO problem, recent work suggested enriching standard random forest (RF) performance regression models with a weighted average of algorithms' performance on training problems that are considered similar to a test problem. More precisely, in this RF+clust approach, the weights are chosen proportionally to the distances of the problems in some feature space. Here in this work, we extend the RF+clust approach by adjusting the distance-based weights with the importance of the features for performance regression. That is, instead of considering cosine distance in the feature space, we consider a weighted distance measure, with weights depending on the relevance of the feature for the regression model. Our empirical evaluation of the modified RF+clust approach on the CEC 2014 benchmark suite confirms its advantages over the naive distance measure. However, we also observe room for improvement, in particular with respect to more expressive feature portfolios. 
\end{abstract}


\section{Introduction}
In black-box optimization, supervised machine learning (ML) models are commonly used for automated algorithm selection~\cite{KerschkeHNT19survey}. The models use representations of the optimization problems in terms of exploratory landscape analysis (ELA) features~\cite{mersmann2011exploratory} to predict the algorithm performance on the problems. Typically, regression models are used~\cite{KerschkeT19AS}. While promising results have been achieved, the models may not make accurate predictions when they are trained on problems that are not representative of the new problems, for which the best-performing algorithm shall be selected.

In the majority of previous works, the predictive power of the ML models is evaluated in leave-one-instance-out (LOIO) scenarios~\cite{jankovic2020landscape,jankovic2021impact,jankovic2021towards}, where different instances from the same problem are present in both, training and test data. The problem instances are obtained with transformations of the same base problem class by using shifting, scaling, and/or rotation~\cite{vskvorc2020understanding}. In such an evaluation scenario there is a guarantee that the training data covers the feature space also of the unseen test instances. As a result, the learned predictive model performs well on new problems. 

The main challenge arises when the evaluation of the predictive model is performed in a leave-one-problem-out (LOPO) manner. That is, all the instances of the same base problem are left out for testing, and no instances of the problem are present during the training of the model. Recent studies~\cite{vskvorc2022transfer,kostovska2022per} show that it is difficult to generalize a predictive model for automated algorithm performance prediction trained on problems from one benchmark suite to problems from another benchmark suite. 

The LOPO performance prediction can be stated as an ML task known as zero-shot learning~\cite{reis2018hyper,larochelle2008zero,wang2019survey}. 
Nikolikj et al. introduced RF+clust for LOPO algorithm performance prediction~\cite{nikolikj2023evoapp}. The approach calibrates the prediction obtained by a Random Forest (RF) model~\cite{biau2016random} for a given test problem with a weighted mean of the algorithm performance on problems from the training data that are the most similar to the test problem, based on their feature representation. The similarity between problems is measured using cosine similarity~\cite{singhal2001modern}. While RF+clust produces promising results, it struggles when problems have very similar feature representations but there is a large difference in algorithm performance on them. In addition, treating all features equally when calculating the similarity between the problem representations can be a weak point of the approach, resulting in similar problems from the training data which are not similar in reality to the test problem.

\textbf{Our contribution:} In this study, we analyze how using a weighted similarity measure to find similar problem instances affects the performance of RF+clust. We evaluate the performance of two weighted RF+clust variants that incorporate feature importance in the similarity measure as weights. We tested two methods for determining the feature importance: an \textit{unsupervised method}, based on clustering, and a \textit{supervised method} based on feature permutation. The results obtained on the CEC 2014 benchmark suite show that using a weighted cosine similarity measure with feature importance as weights can improve the performance of the RF+clust approach. Most of the problems for which new similar problems have been found with a weighted similarity measure and the algorithm performance on them is similar yield better results. Previously for several problems for which similar problems are not found, we can now identify them through the weighted similar measure and improve the prediction. For some problems the approach leads to worse predictions, indicating that the most important features are not expressive enough to distinguish between these problems, with significantly different algorithm behavior. Comparing both weighted variants, it follows that the one that uses the permutation feature importance provides close to the uniform contribution of the features. This is the reason it has more similar prediction results to the standard RF+clust approach.

\textbf{Outline:} The rest of the paper is organized as follows: Section~\ref{sec:realtedwork} presents the overview of the related work, Section~\ref{sec:methodology} introduce the two variants of RF+clust approach, the experimental design is explained in Section~\ref{sec:experiments}. Section~\ref{sec: results} presents the results with discussion, and finally, the conclusions are presented in Section~\ref{sec:conclusion}.

\section{Related work}
\vspace{-1.5mm}

\label{sec:realtedwork}
\subsection{Common ML approaches for automated algorithm performance prediction}
In the black-box optimization context, most of the studies performed in the direction of supervised automated algorithm performance prediction use Exploratory Landscape Analysis (ELA)~\cite{mersmann2011exploratory} for calculation of the feature representation of problems. The ELA features are used as input data for the ML models. The ELA features are calculated by using mathematical and statistical techniques on a set of candidate solutions sampled from the problem decision space. Further, they are combined with different supervised ML models (e.g., Random Forest, XGBoost, Neural Networks, etc.) to predict the performance of an algorithm~\cite{KerschkeT19AS,jankovic2021impact,trajanov2022explainable}. However, the evaluation of the predictive models in almost all studies on this topic is evaluated in the LOIO scenario.

Few studies that have been published in 2022, analyze models' generalization power in the LOPO scenario. \v{S}kvorc et al.~\cite{vskvorc2022transfer} showed that a predictive model has lower generalization in such evaluation scenario. They have analyzed this on the real-parameter black-box optimization benchmarking (BBOB) benchmark suite~\cite{hansen2010real}. In addition, they have shown that a model trained on the BBOB benchmark suite provides poor predictive results when it is used to predict the performance of a benchmark suite of artificially generated problems. Kostovska et al. have analyzed a ``per-run'' algorithm selection scheme~\cite{kostovska2022per} by using trajectory-calculated ELA features (i.e., the samples for calculating them are the samples observed by the algorithm during its run) as input data to predict the performance. They have shown that a model learned on the BBOB benchmark suite has poor predictive results for the Nevergad benchmark suite~\cite{nevergrad} and vice versa.

\subsection{RF+clust for leave-one-problem-out (LOPO) performance prediction}
RF+clust is a recently proposed approach for LOPO algorithm performance prediction. The idea behind the approach is to improve the predictions of a standard supervised model when the test problem landscape representation is not present in the training set. The approach consists of the following three steps:
\begin{itemize}
    \item Training a regression model on a set of training problems represented by their landscape features. As the name suggests, the RF+clust approach uses random forest (RF) regression models to obtain the predicted algorithm performance $\widehat{y}_q$ for the test problem.
    
    \item The second step consists of setting a predefined similarity threshold for problem similarity. Based on the feature representation of the problems, the $k$-nearest problems from the training set whose similarity $s$ with the test problem is greater or equal to the predefined threshold are selected. The number $k$ can be different for different problems and it depends on the predefined threshold. The approach uses cosine similarity as a similarity measure. For the selected problems, the actual performance of the algorithm on them is retrieved, $y_1,y_2,\dots,y_k$.
    
    \item The last step is calibrating the RF prediction $\widehat{y}_q$ with the performance of the algorithm on the selected training problems from the previous step and obtaining the final prediction for the test problem. This is performed by the following aggregation: 
    $\widehat{y}_{q, \text{final}}$ 
    = $\left(\widehat{y}_{q}^i + F(y_{1}, y_{2},\ldots, y_{k})\right)/2$, where 
    $F(\dots)$ = $\sum_{i=1}^{k} w_{i}y_{i}$. 
    The weight indicates how much each of the selected problems contributes to the calibration and it is calculated based on its similarity to the test problem, 
    $w_{i}=s_{i}/\sum_{i=1}^k s_i$. 
    In cases where there are no similar problems for the predefined threshold from the training set, the prediction is only based on the RF prediction 
    $\widehat{y}_{q, \text{final}}$ = $\widehat{y}_q$.
\end{itemize}

\section{Sensitivity analysis of RF+clust for LOPO performance prediction}
\label{sec:methodology}
In this paper, we perform a sensitivity analysis of RF+clust, where the main difference with the original RF+clust, is to learn the importance of each feature and further incorporate it in the procedure for finding the $k$-nearest problems from the training set, that are used to calibrate the prediction obtained by the RF model on the test set. The main motivation behind this is, the more important features to have more influence in the similarity score. Let us assume that we have $p$ features. 

Here, we use and compare two different methods for learning feature importance, one unsupervised (proposed by us) and one supervised (that is a well-established approach). Details about the approaches are provided below:
\begin{itemize}
    \item Unsupervised learning - the problems from the training set are clustered with the full feature portfolio into $m$ clusters. Further, the same clustering is performed $p$ times, each time removing one feature and clustering the problems using the $p-1$ features into the same number of previous estimated $m$ clusters. At the same time, we are counting on how many problems the obtained clusters with the full portfolio differ from the new clusters. The cumulative number of problems in which the clustering differs ($n_{\text{diff}}$) is used to calculate the weight of the omitted feature. A larger number of problems indicates higher importance since omitting the feature places the problems all around the problem landscape space, while a smaller number indicates that this feature is not important in distributing the problems. After we obtain the number of problems that change their placement in the problem landscape space for each feature, we calculate their weight by using $w_i={n_{\text{diff}}}_i/\sum_{i=1}^{p}{n_{\text{diff}}}_i$.
    \item Permutation feature importance - The permutation feature importance, $\text{perm}$, is defined to be the amount by which a model's performance drops when single feature values are randomly shuffled~\cite{breiman2001random}. The main notion behind this procedure is that it breaks the relationship between the feature and the target, thus the drop in the model performance is an indication of how much the model depends on the feature. This technique benefits from being model agnostic and usually, the final feature importance is calculated by shuffling the feature multiple times with different permutations of the feature. The weights are then calculated as $w_i={\text{perm}}_i/\sum_{i=1}^{p}{\text{perm}}_i$.
\end{itemize}

After obtaining the weights for each feature, the similarity of the test problem with the training problems is calculated using a weighted cosine similarity
\begin{equation}
    \text{cosine}(u,v)=\frac{\sum_{i=1}^p w_i^2 u_iv_i}{\sqrt{\sum_{i=1}^p (w_iu_i)^2}\sqrt{\sum_{i=1}^p (w_iv_i)^2}},
\end{equation}
where $u=(u_1,\dots, u_p)$ and $v=(v_1,\dots, v_p)$ are the feature representations of the problems with $p$ ELA features.

The $k$-nearest problems that are selected by applying the similarity threshold to the weighted cosine similarity measure, are the ones used to calibrate the RF prediction.

\section{Experimental design}
\label{sec:experiments}
For better comparison, we build our experiments on the same data as~\cite{nikolikj2023evoapp}.

\textbf{Problem portfolio.} The proposed approach is evaluated on the 2014 CEC Special Sessions \& Competitions (CEC 2014) benchmark suite~\cite{liang2013problem}. It consists of 30 benchmark problems. The problem dimension $D$ is set to 10.

\begin{figure}[!htb]
\centering
\includegraphics[width=0.5\textwidth]{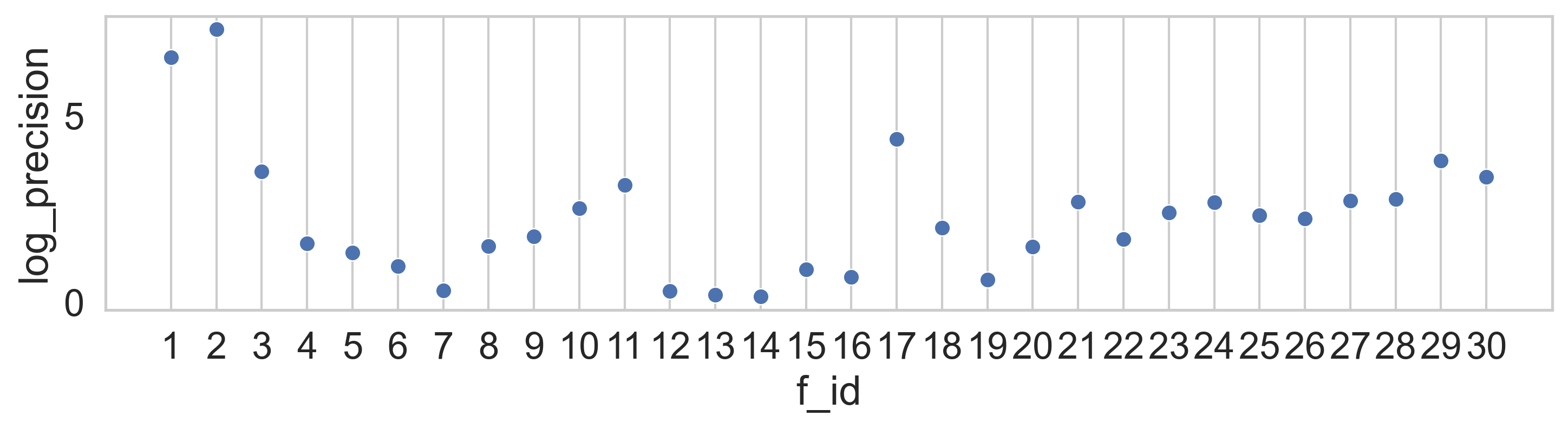}
\caption{Best solution precision (log-scale) obtained by DE$1$, for a budget of 5000 function evaluations, per problem instance in the CEC 2014 benchmark suite.}
\label{raw_performance}
\end{figure}

\textbf{Algorithm portfolio.} Three randomly selected Differential Evolution (DE) configurations are included in the analysis. Their hyper-parameters are set as presented in~\cite{nikolikj2023evoapp}. The population size of the algorithm is set equal to the problem dimension (10). Each configuration is run 30 times on each CEC problem and the precision after a budget of $500D=5000$ function evaluations has been stored. Finally, we report the median precision over all 30 runs for each pair of an algorithm configuration and a CEC problem. In the ML task, we consider the logarithm ($\log_{10}$) of the median precision as a target for prediction. Figure~\ref{raw_performance} presents $DE$1 performance (log-scale) obtained per benchmark problem on the CEC 2014 benchmark suite. The random selection of the DE configurations is because we focus on presenting the utility of the methodology, which is a method that can be used for any choice of algorithms and their hyperparameters.

\textbf{Exploratory Landscape Analysis (ELA).} To extract features that describe the properties of each CEC problem, we utilize the ELA technique which is the most commonly used meta-representation for continuous single-objective optimization problems~\cite{mersmann2011exploratory}. 64 ELA features are taken from a previous study~\cite{lang2021exploratory}, where they are calculated using Improved Latin Hypercube Sampling (ILHS)~\cite{xu2017improved} with a sample size of 800$D$ (8000) and 30 runs. The large sample size can be at a high cost, however, this was performed to reduce the randomness in the feature extraction process. 

For the final analysis, as a feature portfolio, we have selected only the uncorrelated features. Pearson correlation coefficient~\cite{cohen2009pearson} of 0.9 is used as a threshold to retrieve the highly correlated feature pairs. Next, the correlated features are divided into groups of correlated features, where all the features in the group have been highly correlated with each other. This problem can be translated into a graph problem where all the features are nodes and pairs of features satisfying correlation $>$ 0.9 are edges. The task of finding all ``correlated groups" translates into finding all complete sub-graphs in the graph with more than 2 nodes. The implementation is done with the Python package \textit{NetworkX v.2.8.4}~\cite{hagberg2008exploring}. Then the RF model with default parameters was evaluated against every single feature from the group, the feature that resulted with the lowest mean absolute error (MAE) was chosen to be kept and the others were discarded. We need to point out here that the selected feature portfolio is different for different algorithms and even for different folds of the same algorithm.

Table~\ref{t:ml_performance} shows the ML model performance aggregated over all folds, when all features available have been used and when only the uncorrelated features (around 30 depending on the fold) have been used. Comparing the train and test errors for the different feature portfolios, we can see that the performance is only slightly degraded for the uncorrelated feature portfolio, for all the algorithms. For further experiments, we have selected the uncorrelated feature set. Performing this selection we have reduced the risk of overfitting the prediction models. 

\begin{table}[!ht]
\caption{Mean absolute error (MAE) obtained by the RF models when predicting the performance of the three DE configurations, using all and the uncorrelated features. The values in the table represent the MAE over 30 folds.}
\label{t:ml_performance}
\centering
\begin{tabular}{ccccccc}
\hline
features &  algorithm & MAE\_train &  MAE\_test \\
\hline
uncorrelated  &  DE1 & 0.448384 &  1.267805 \\
all      &  DE1 & 0.453909 &  1.208038 \\
uncorrelated  &  DE2 & 0.384077 &  1.077129 \\
all      &  DE2 & 0.392174 &  1.054833 \\
uncorrelated  &  DE3 &  0.375334 &  1.023077 \\
all      &   DE3 & 0.376666 &  1.018123 \\
\hline
\end{tabular}
\end{table}

\begin{figure}[!htb]
\centering
\includegraphics[width=0.5\textwidth]{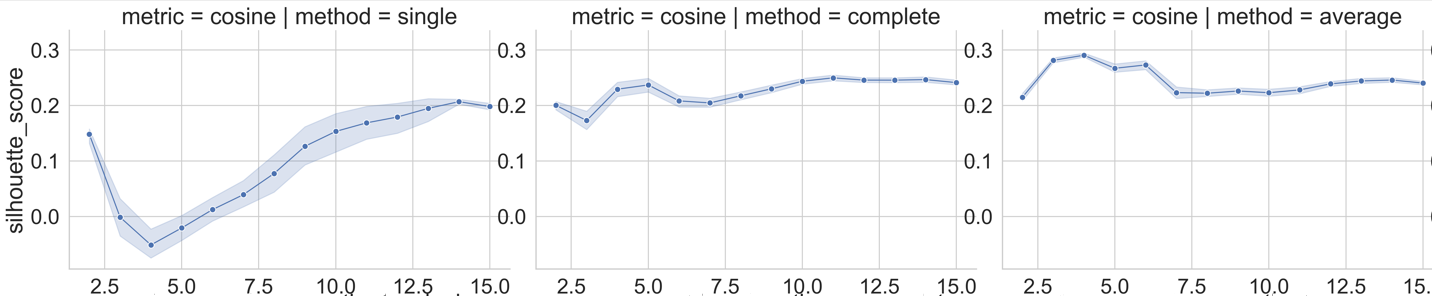}
\caption{Estimating the number of clusters and tuning of clustering hyper-parameters for algorithm DE1.}
\label{figclust}
\end{figure}

\textbf{Feature importance.}
To learn the weights of the features required to perform the sensitivity analysis of the RF+clust approach, we use i) hierarchical clustering and ii) permutation feature importance.  Both methods have been selected to test different variants of RF+clust methodology. In the case of the hierarchical clustering, the number of clusters together with the selected hyper-parameters has been estimated using hierarchical clustering~\cite{murtagh2012algorithms} which is often the first option for very small data sets such as the 29 training problems used in this case. Sub-figures in Figure~\ref{figclust} show clustering performance (y-axis) with standard deviation over folds, for different numbers of clusters (x-axis), when using different parameters for the clustering algorithm. The implementation has been done using the \textit{scipy v.1.9.3} Python package with $metric$ set to cosine similarity as a distance measure, $method$ set to ``average" and the number of clusters $m$ set to 4. For the permutation feature importance, the implementation has been done using the \textit{scikit-learn v.1.0.2} package in Python by setting the number of permutations, $n\_repeats$, to 15 and $random\_state$ to one for reproducibility of the results.

\textbf{RF+clust model training and evaluation.} We follow the previously introduced RF+clust variant, where the RF model (from the \textit{scikit-learn} package in Python) is trained in an STR learning scenario for each algorithm configuration separately. The evaluation is performed in the LOPO scenario (i.e., 29 problems are used for training and one for testing), where the prediction errors are the absolute distances of the prediction of the precision to the true precision value of the algorithm. Since we have 30 problems, the learning process is repeated 30 times, each time one problem is out, so all steps are repeated including feature importance learning and training a regression model only on the training set.

\section{Results and discussion}
\label{sec: results}
We apply the approach to three random DE configurations on the CEC 2014 benchmark suite. Due to space limitations, we present here some selected results for algorithm DE$1$ in more detail, while for the other results, similar findings were noticed and are available at~\cite{gitRFClust}.

\subsection{Sensitivity analysis}

\begin{figure}[!htb]
\centering
\includegraphics[width=0.47\textwidth]{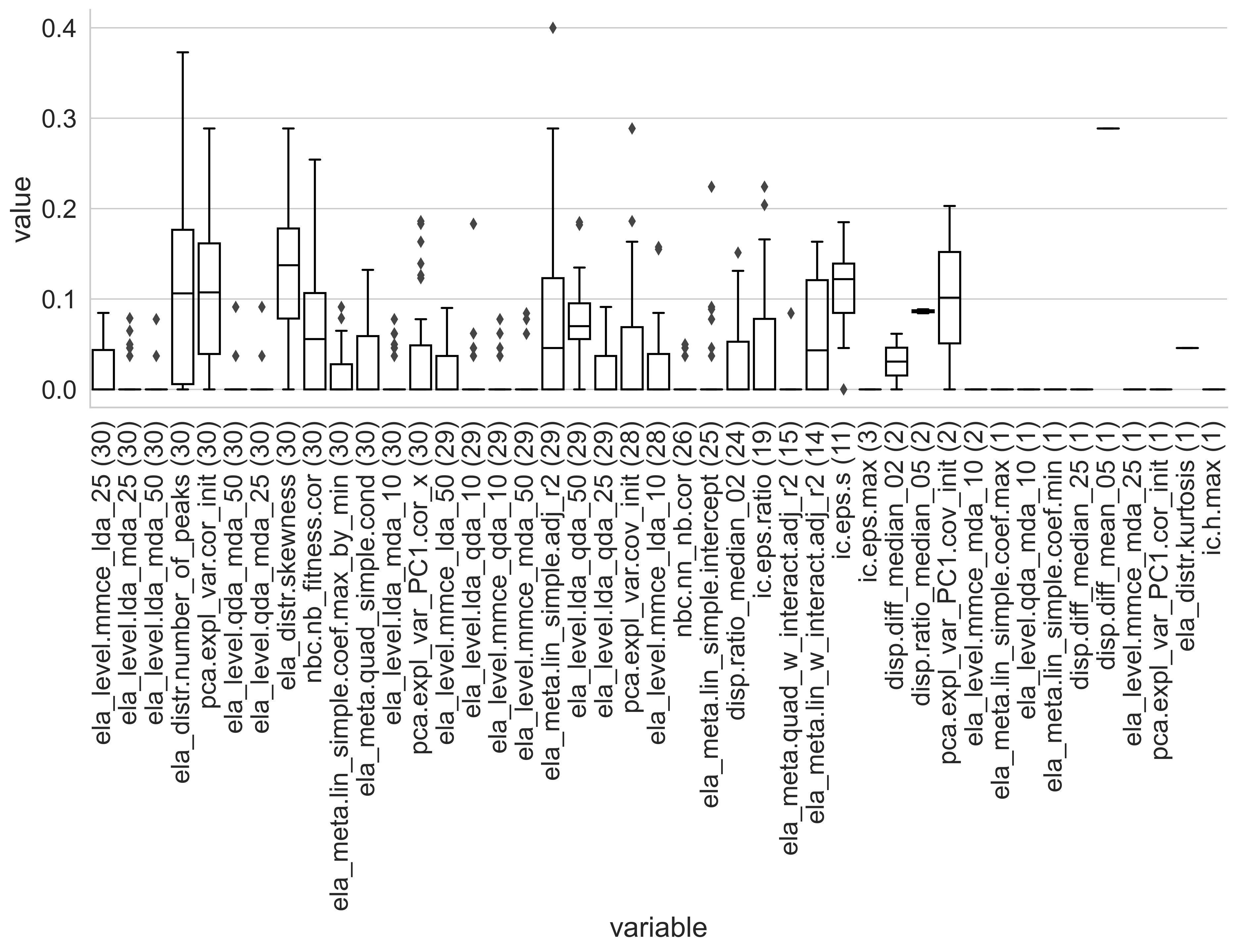}
\caption{Feature weights box-plot over all folds for algorithm DE$1$ obtained with the unsupervised feature importance. The numbers in brackets indicate in how many folds the feature was selected in the feature selection process.}
\label{fig_DE1_weights_clus}
\end{figure}

Figure~\ref{fig_DE1_weights_clus} presents the box plots of the distribution of the weights for each of the features obtained for DE$1$ across all 30 folds. The weights are calculated with the unsupervised feature importance approach. On the x-axis, we have all the selected features and for each one in brackets, we present in how many folds it has been selected as an uncorrelated feature. The y-axis (i.e., value) represents the calculated weight. The figure shows that half of the features selected in the uncorrelated feature portfolio for each fold are not important since they have weights equal to zero in almost all of the folds, which means that they are not used to find similar problems used for the calibration. 

\begin{figure}[!htb]
\centering
\includegraphics[width=0.47\textwidth]{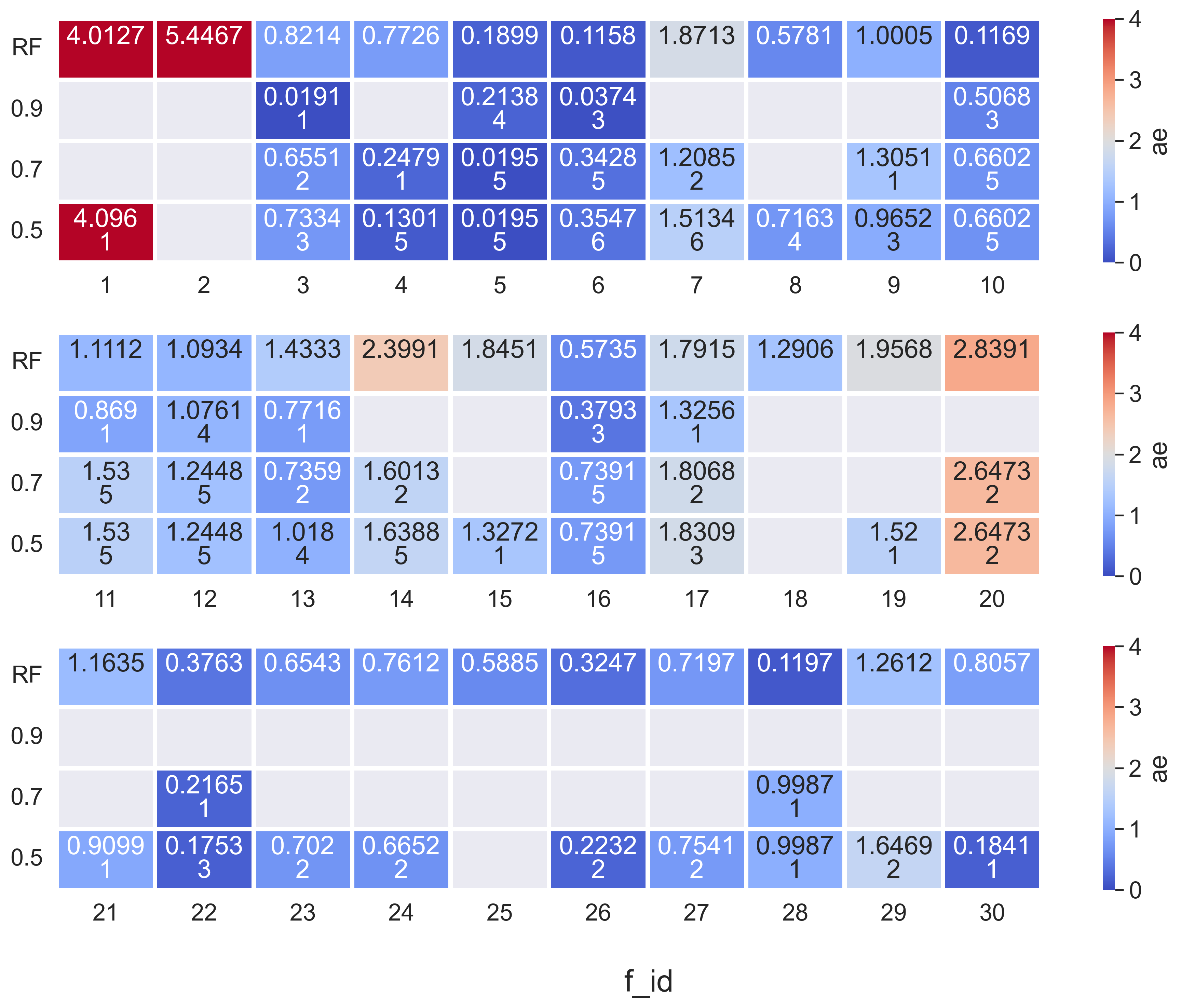}
\vspace{-2mm}
\caption{Error comparison between RF and RF+clust (with \textbf{cosine similarity} and similarity threshold of 0.5, 0.7, and 0.9) in predicting the performance of DE$1$ for each problem instance in the CEC 2014 suite.}
\label{RFclust_error_none}
\end{figure}

Figure~\ref{RFclust_error_none} presents the comparison of prediction errors between RF and the initial RF+clust approach for similarity thresholds of 0.5, 0.7, and 0.9 in a LOPO scenario. The first row shows the errors obtained by a standard RF model trained in the LOPO scenario. Each cell of the heatmap represents the mean absolute error obtained by the models on the test set. The numbers under the model error indicate the number of similar problems above the corresponding threshold that have been selected from the training set for the calibration of the prediction. The blank cells in the heatmap are problems for which RF+clust provides the same result as the standard RF model because for those problems we could not find similar problems from the training data to calibrate the prediction. The column names in the heatmap presented below are the problems from the CEC 2014 suite (i.e., f$\_$id).

\begin{figure}[!htb]
\centering
\includegraphics[width=0.47\textwidth]{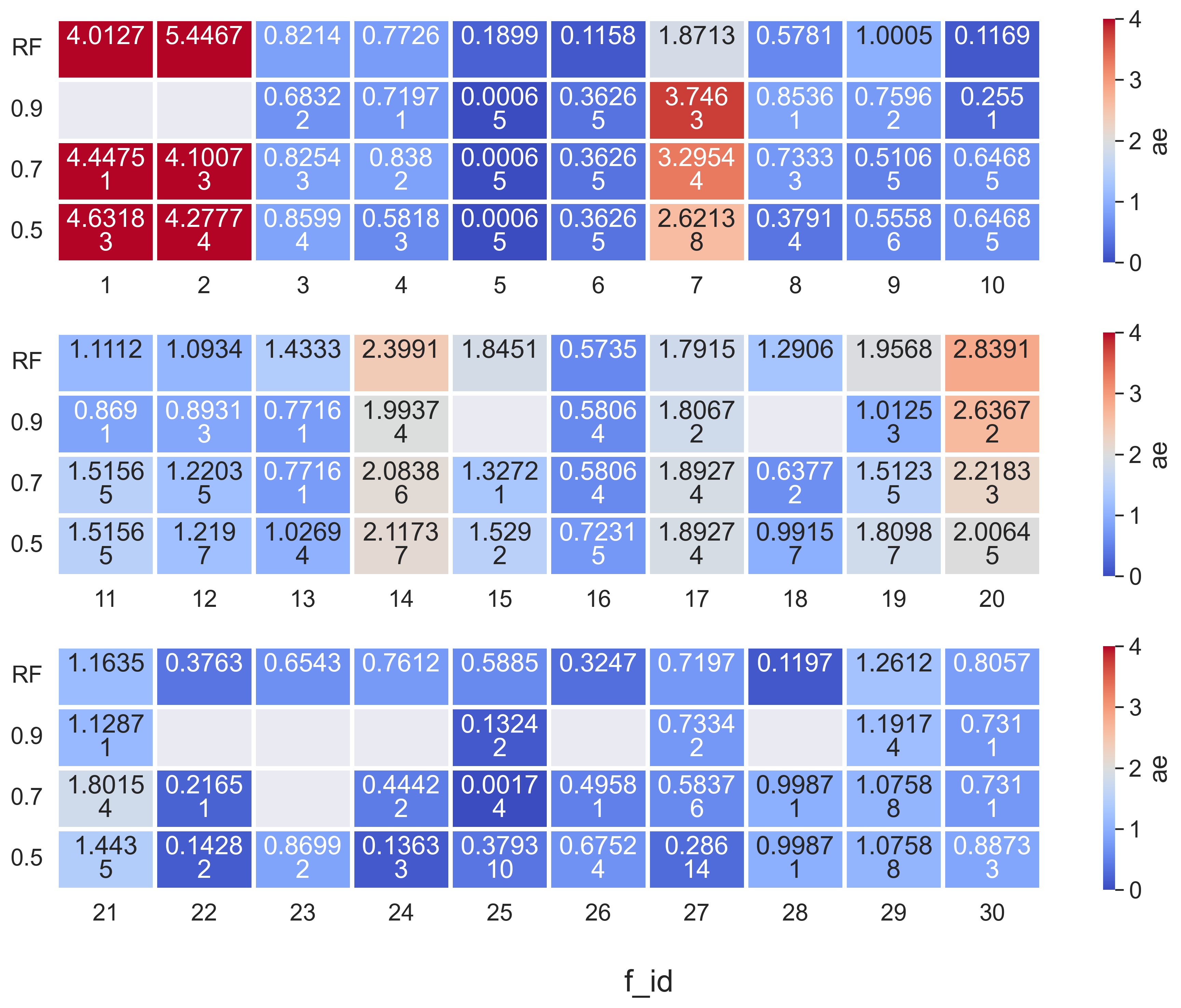}
\vspace{-2mm}
\caption{Error comparison between RF and RF+clust (with
\textbf{weighted cosine similarity with weights calculated by the
unsupervised approach for feature importance} and similarity threshold of 0.5, 0.7, and 0.9) in predicting the performance of DE$1$ for each problem
instance in the CEC 2014 suite}
\label{RFclust_error_unsupervised}
\end{figure}

\begin{figure}[!htb]
\centering
\includegraphics[width=0.47\textwidth]{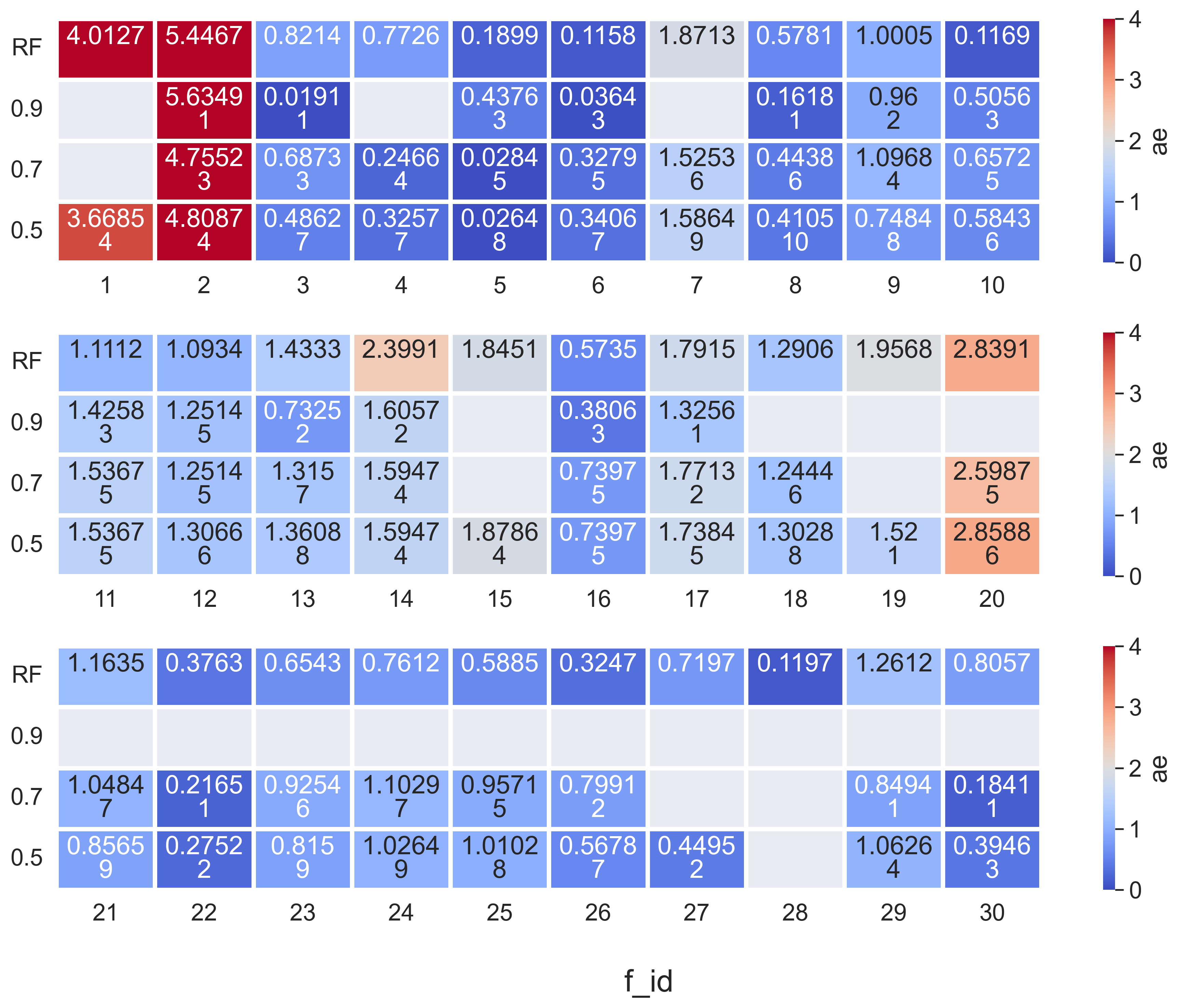}
\vspace{-3mm}
\caption{Error comparison between RF and RF+clust (with
\textbf{weighted cosine similarity with weights calculated by the
with weights calculated by the permutation feature importance} and similarity threshold of 0.5, 0.7, and 0.9) in predicting the performance
of DE$1$ for each problem instance in the CEC 2014 suite.}
\label{RFclust_error_permutation}
\end{figure}

Figures~\ref{RFclust_error_unsupervised} and~\ref{RFclust_error_permutation} demonstrate the results obtained with the weighted RF+clust approach which uses the unsupervised feature importance or permutation feature importance respectively as feature weights to find similar problems used for calibration. The weighted approach calculates the weighted cosine similarity of the problem representations. We can notice in both figures that the similarity between the problems can be influenced by the weighting, as the number of similar problems changes in many cases as compared to Figure~\ref{RFclust_error_none}.

\subsection{In-depth analysis}
\begin{figure}
\begin{subfigure}[b]{4cm}
  \centering
  \includegraphics[width=4cm]{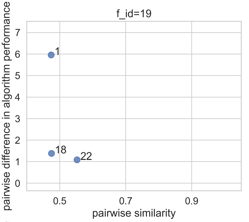}  
  \caption{cosine similarity}
  \label{fig:RF+clust_improvement_1}
\end{subfigure}
\hfill
\begin{subfigure}[b]{4cm}
  \centering
  \includegraphics[width=4cm]{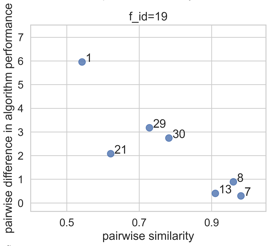}  
  \caption{weighted}
  \label{fig:RF+clust_improvement_2}
\end{subfigure}

\begin{subfigure}[b]{4cm}
  \centering
  \includegraphics[width=4cm]{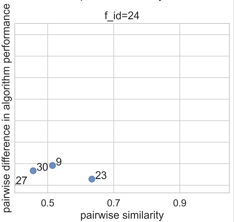}  
  \caption{cosine similarity}
  \label{fig:RF+clust_improvement_3}
\end{subfigure}
\hfill
\begin{subfigure}[b]{4cm}
  \centering
  \includegraphics[width=4cm]{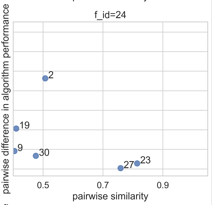}  
  \caption{weighted}
  \label{fig:RF+clust_improvement_4}
\end{subfigure}

\caption{The relationship between the pairwise cosine and weighted cosine similarity by unsupervised feature importance of the feature representations (x-axis) and the difference in DE$1$ performance (y-axis), for the \textbf{19th} and \textbf{24th} problem accordingly, with other problems in the CEC 2014 suite.}
\label{fig:RF+clust_improvement}
\end{figure}

The heatmap of the weighted RF+clust approach  with weights learned by the unsupervised feature importance (see Figure~\ref{RFclust_error_unsupervised}), shows lower errors for the following problems: 2, 5, 9, 10, 12, 19, and 24, for all similarity thresholds compared to the initial RF+clust approach. To provide an explanation of why this happens, the 19th problem is analyzed in more detail. With the initial RF+clust, there are no similar problems found above similarity thresholds of 0.7 and 0.9 (see Figure~\ref{RFclust_error_none}). With the weighted RF+clust, we can detect three similar problems from the training data above $>$ 0.9. The result is visible in more detail in Figures~\ref{fig:RF+clust_improvement_1} and~\ref{fig:RF+clust_improvement_2}, which show the relationship between the pairwise similarity of the ELA features representation (x-axis) and the absolute pairwise difference in the (ground truth) performance of the optimization algorithm (y-axis) of the 19th problem (as indicated in the plot's title) with the other problems. 
Figure~\ref{fig:RF+clust_improvement_1} presents results of using the cosine similarity between the ELA representations, while Figure~\ref{fig:RF+clust_improvement_2} uses the weighted cosine similarity.
Comparing the two plots, the left one shows the initial RF+clust results where it is visible that there are no similar problems with similarity above 0.9, and the right one shows the weighted RF+clust results from which it is visible that three problems have been found as similar with a weighted cosine similarity of above 0.9. In addition, we can see that the difference in ground truth performance of the algorithm on the 19th problem and the three selected similar problems is low. The algorithm has similar behavior on these problems in reality (see also Figure~\ref{raw_performance}), and using them for the calibration helps to obtain lower predictive errors. The same conclusion can be drawn for the 24th problem presented in Figures~\ref{fig:RF+clust_improvement_3} and~\ref{fig:RF+clust_improvement_4} with a similarity threshold of 0.7. This result indicates that using the weighted approach can help us to identify and use problems that are more effective in the calibration step. 

\begin{figure}
\begin{subfigure}[b]{4cm}
  \centering
  \includegraphics[width=4cm]{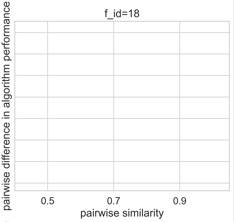}  
  \caption{cosine similarity}
  \label{fig:RF+clust-0_similar-1}
\end{subfigure}
\hfill
\begin{subfigure}[b]{4cm}
  \centering
  \includegraphics[width=4cm]{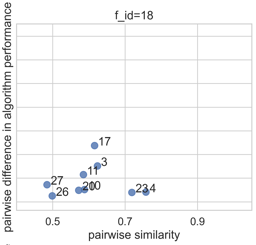}  
  \caption{weighted}
  \label{fig:RF+clust-0_similar-2}
\end{subfigure}

\begin{subfigure}[b]{4cm}
  \centering
  \includegraphics[width=4cm]{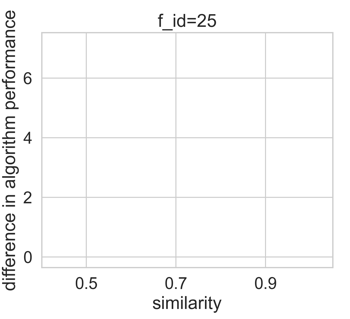}  
  \caption{cosine similarity}
  \label{fig:RF+clust-0_similar-3}
\end{subfigure}
\hfill
\begin{subfigure}[b]{4cm}
  \centering
  \includegraphics[width=4cm]{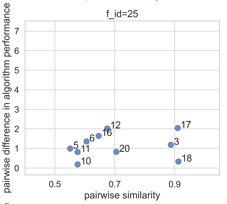}  
  \caption{weighted}
  \label{fig:RF+clust-0_similar-4}
\end{subfigure}

\caption{The relationship between the pairwise cosine and weighted cosine similarity by unsupervised feature importance of the feature representations (x-axis) and the difference in DE$1$ performance (y-axis), for the \textbf{18th} and \textbf{25th} problem accordingly, with other problems in the CEC 2014 suite.}
\label{fig:RF+clust-0_similar}
\end{figure}

There are also problems such as the 18th and 25th for which there are no similar problems found using cosine similarity (see Figure~\ref{fig:RF+clust-0_similar-1} and ~\ref{fig:RF+clust-0_similar-3} accordingly). In this case, the initial RF+clust has the same prediction error as the standard RF model for all similarity thresholds (0.5, 0.7, 0.9). By applying the weighted RF+clust with the unsupervised feature importance, the 18th and 25th problems are brought closer to some similar problems from the training data in the feature space as demonstrated in Figures~\ref{fig:RF+clust-0_similar-2} and~\ref{fig:RF+clust-0_similar-4}. This helps to reduce the model error for these problems as visible in Figure~\ref{RFclust_error_unsupervised}.

For the 20th problem and similarity threshold 0.5 we seem to detect five similar problems (see Figure~\ref{RFclust_error_unsupervised}) with weighted cosine similarity and improve the model performance prediction quite a lot, compared to the case when weights are not used and only two problems are there (see Figure~\ref{RFclust_error_none}). From the figures, it is visible that with the weighted approach, three problems (the 15th, 18th, and 22nd) enriched the previous two (the 3rd and 17th) which helped the calibration process. The algorithm has very similar behavior on three problems that are brought closer with the weighted approach, as on the 20th problem. However, on the remaining two problems (3rd and 17th), we can see that even with high similarity in the landscape space, the difference in algorithm performance is larger in reality, so using their performance to calibrate the prediction yields a larger error. This indicates that there are problems for which even the most important ELA features are not expressive enough (i.e., very similar ELA landscape representation but different algorithm performance).

\begin{figure}
\begin{subfigure}[b]{4cm}
  \centering
  \includegraphics[width=4cm]{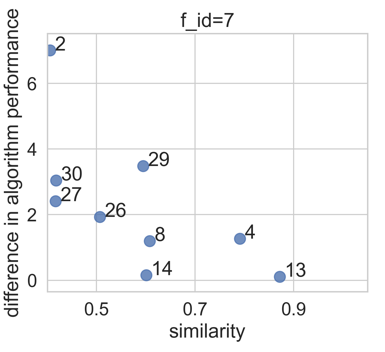}  
  \caption{cosine similarity}
  \label{fig:RF+clust-worse-1}
\end{subfigure}
\hfill
\begin{subfigure}[b]{4cm}
  \centering
  \includegraphics[width=4cm]{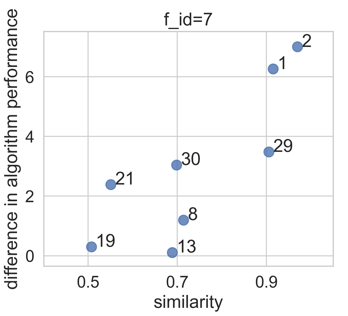}  
  \caption{weighted}
  \label{fig:RF+clust-worse-2}
\end{subfigure}

\begin{subfigure}[b]{4cm}
  \centering
  \includegraphics[width=4cm]{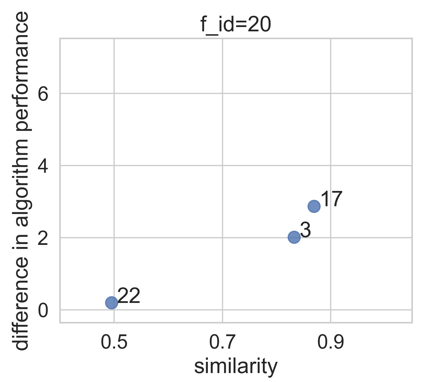}  
  \caption{cosine similarity}
  \label{fig:RF+clust-worse-3}
\end{subfigure}
\hfill
\begin{subfigure}[b]{4cm}
  \centering
  \includegraphics[width=4cm]{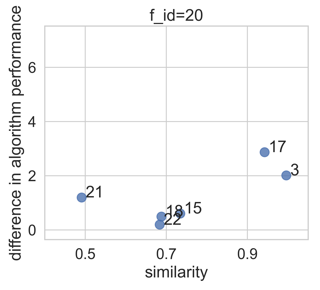}  
  \caption{weighted}
  \label{fig:RF+clust-worse-4}
\end{subfigure}

\caption{The relationship between the pairwise a) cosine and b) weighted cosine similarity by unsupervised feature importance of the feature representations (x-axis) and the difference in DE$1$ performance (y-axis), for the \textbf{7th} and \textbf{20th} problem accordingly, with other problems in the CEC 2014 suite.}
\label{fig:RF+clust-}
\end{figure}

Figures~\ref{fig:RF+clust-worse-1} and~\ref{fig:RF+clust-worse-2} show another downside of using feature importance as weights for finding similar problems. In cases when all the features have an equal contribution to the cosine similarity, RF+clust provides better prediction than a standard RF model. Here, it is visible that two problems (the 4th and the 13th) help the calibration for a similarity threshold over 0.7, for which the performance of the algorithm is similar to the performance achieved on the 7th problem (see Figure~\ref{RFclust_error_none}). In the case of the weighted variant of RF+clust, the prediction is worse even than a standard RF prediction. This happens since the learned weights by the unsupervised approach for feature importance brought a lot of similar problems (the 1st, 2nd, 8th, 30th) with similarity over 0.7, however, the difference in the performance of the algorithm on those problems with the performance achieved on the 7th problem is higher. This result again points out that similar landscape representation may not always be a guarantee of similar performance, which opens a new research direction of inventing new more robust problem representations that will catch the relation between the feature landscape space and the performance space of the algorithm.

\begin{figure}[!htb]
\centering
\includegraphics[width=0.47\textwidth]{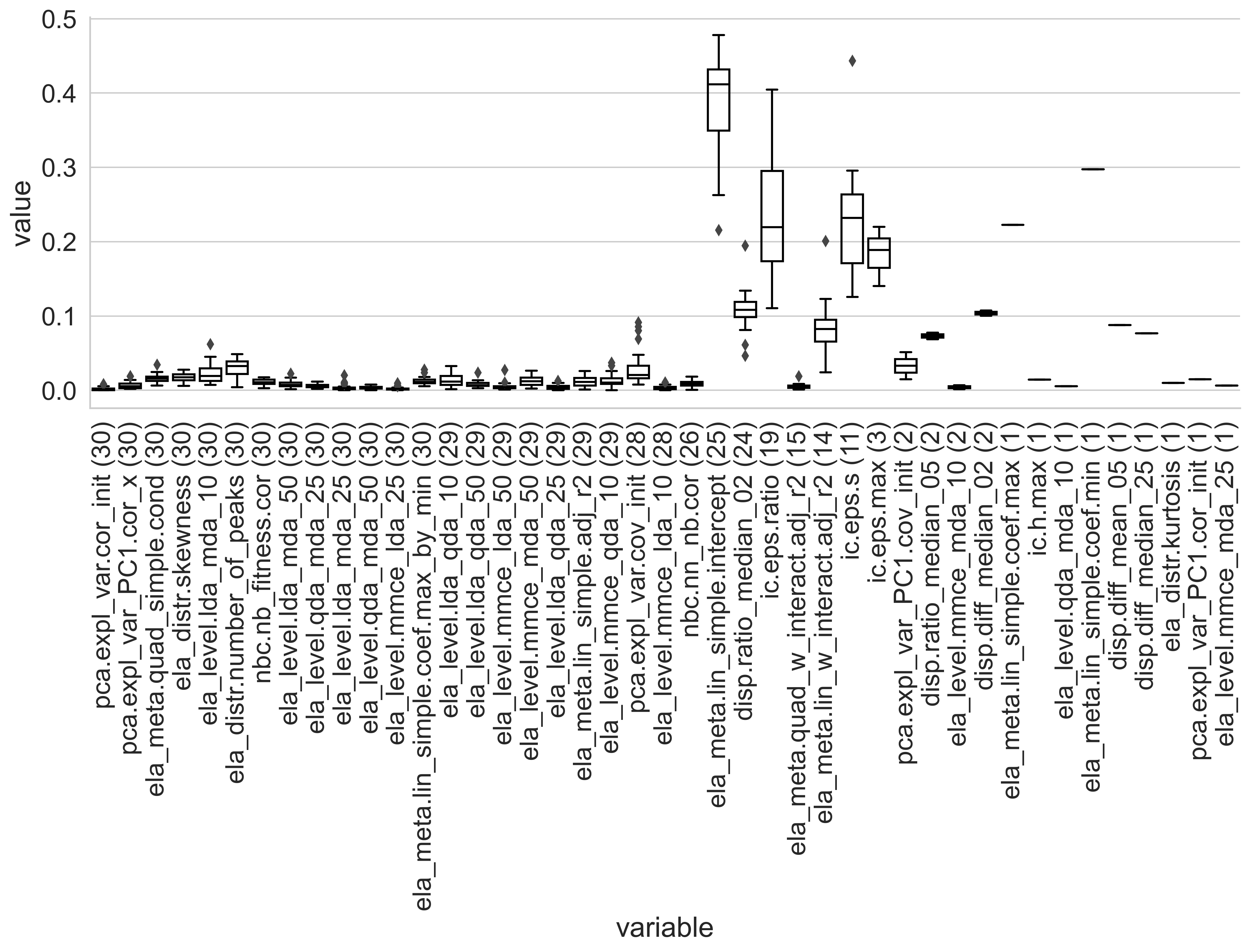}
\caption{Feature weights box-plot over all folds for algorithm DE$1$ obtained with permutation feature importance. The numbers in brackets indicate in how many folds the feature was selected in the feature selection process.}
\label{fig_DE1_weights_perm}
\end{figure}

Figure~\ref{RFclust_error_permutation} presents the results of the RF+clust variant which uses the permutation importance as weights. We can see that using these weights also changes the number of similar problems retrieved, however the results are similar with the initial RF+clust, with slight changes. Analyzing the weights that were obtained by this approach it seems that they are close to uniform for most of the features, with only a few features showing bigger importance as shown in Figure~\ref{fig_DE1_weights_perm}. However, those features are selected as uncorrelated in the feature portfolio only for around half of the folds.

\begin{figure}[!htb]
\centering
\includegraphics[width=0.47\textwidth]{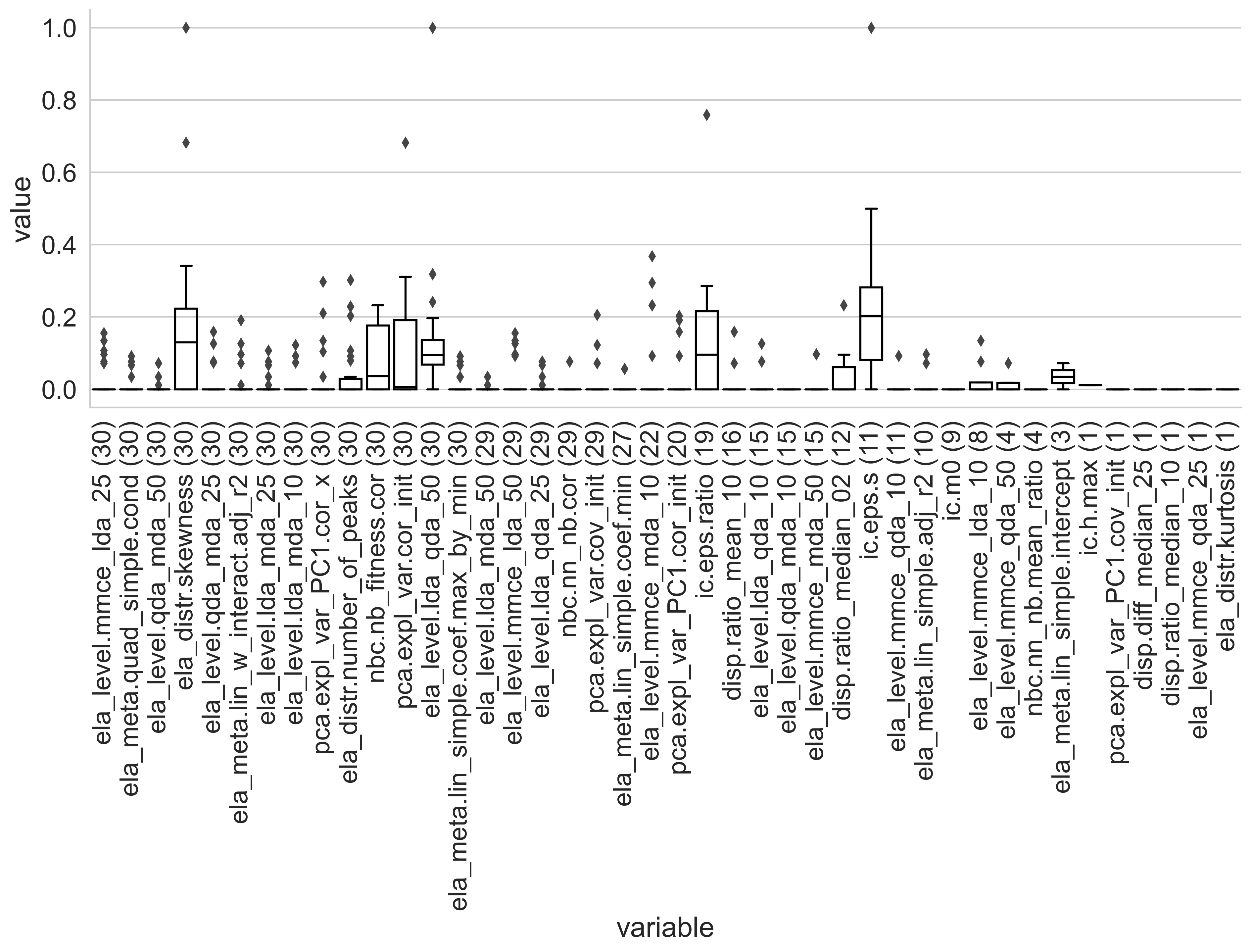}
\caption{Feature weights box-plot over all folds for algorithm DE$3$ obtained by the unsupervised feature importance.}
\label{fig_DE3_unuspervised}
\end{figure}

Table~\ref{tab:mae_general} provides the mean absolute prediction error across all 30 problems for a standard RF model and all variants of the RF+clust model for different similarity threshold values (0.5, 0.7, and 0.9) when they are used to predict the performance of three different DE configurations. The RF+clust approach provides better errors (i.e., bold values in Table~\ref{tab:mae_general}) than a standard RF model. Even if there are small improvements on average, from Figures~\ref{RFclust_error_none},~\ref{RFclust_error_unsupervised},~\ref{RFclust_error_permutation}, it is obvious that for some problems big improvements are obtained. For DE$1$, all variants of RF+clust provide a better result than the standard RF model for all similarity thresholds. For DE$2$, all variants with a similarity of 0.9 provide better prediction results than a standard RF model with the best result achieved when the weights are learned by permutation feature importance. In the case of DE$3$, the best result has been achieved for the variant when all features have the same contribution. Here, it is obvious that the result is similar to the result achieved by a standard RF model and all RF+clust variants and thresholds. To investigate why this happens, Figure~\ref{fig_DE3_unuspervised} presents the distribution of weights for each feature in the case of DE3. From it, we can see that most of the features have weight zero, so when similar problems are searched for we are deciding only based on a few features. From the results, it follows using a 0.9 similarity threshold can provide better results for all DE configurations. The red values reported in Table~\ref{tab:mae_general} are the models with the smallest MAE for predicting each DE configuration.

\begin{table}[!htp]\centering
\caption{MAE for the three DE configurations, where the bold values represent cases when the RF+clust is better than the RF model and the values in red represent the best model.}\label{tab:mae_general}
\scriptsize
\begin{tabular}{lrrrrr}\toprule
&s &DE1 &DE2 &DE3 \\\midrule
RF & &1.267805 &1.077129 &1.023077 \\\midrule
\multirow{3}{*}{RF + clust} &0.9 &\textbf{1.199533} &\textbf{1.042411} &\textbf{\textcolor{red}{1.004657}} \\
&0.7 &\textbf{1.245611} &1.106662 &1.063595 \\
&0.5 &\textbf{1.209003} &1.128300 &1.059814 \\\midrule
\multirow{3}{*}{RF + clust (unsup.)} &0.9 &\textbf{1.223082} &\textbf{1.072002} &1.025545 \\
&0.7 &\textbf{1.217473} &1.120936 &1.078338 \\
&0.5 &\textbf{1.221436} &1.078583 &1.078182 \\\midrule
\multirow{3}{*}{RF + clust (perm.)} &0.9 &\textbf{1.194667} &\textbf{\textcolor{red}{0.971472}} &1.023077 \\
&0.7 &\textbf{1.218578} &1.099617 &1.048660 \\
&0.5 &\textbf{\textcolor{red}{1.180002}} &\textbf{1.052838} &1.106005 \\
\bottomrule
\end{tabular}
\end{table}

\section{Conclusion}
\label{sec:conclusion}
In this study, we performed a sensitivity analysis of RF+clust, a method for leave-one-problem-out (LOPO) performance prediction for black-box optimization algorithms. The main idea behind the RF+clust approach is to calibrate the prediction of a standard RF model with the performance achieved by the algorithm on similar problems  in the training data. In the original RF+clust approach, the similarity between problems is measured by the cosine similarity of the problem landscape features, with all features contributing equally to the similarity measure. For our sensitivity analysis, we tested two new weighted RF+clust variants that use a weighted contribution of each feature to the distance measure. The weights are calculated using a feature importance method. We evaluated two feature-importance approaches: an unsupervised one, based on clustering, and a supervised one, based on permutation. In the future, other feature importance measures can be included in the analysis.

The results performed on the CEC 2014 benchmark suite indicate that RF+clust performance can be further improved by using feature importance as weights. Better results are achieved for problems for which more or new problems (compared to the original RF+clust)  were found similar based on the most important features for which also the algorithm behaves similarly. For problems for which there were no similar problems, we could now successfully find problems with similar feature representations. However, there are also problems for which the proposed approach led to worse prediction results. Such results indicate that even the most important features were not expressive enough to discriminate between these problems on which the algorithm the behavior of the algorithm significantly differs.

Our results open several directions for future research. First, we are going to focus on selecting different feature portfolios for different sets of problems that can lead to robust landscape representation. Next, we are going to test problem representations that are learned by the algorithm behavior and capture the relation between the problem and the performance space. Last, but not least, we are going to test the approach using different ML models. That is, we plan to evaluate the advantage of the RF+clust approach when combined with different regression models (as opposed to the random forest models considered so far).




\end{document}